\documentclass{article}

\usepackage{microtype}
\usepackage{graphicx}
\usepackage{subcaption}
\usepackage{booktabs} 
\usepackage{listings}

\usepackage{hyperref}
\usepackage{lipsum}

\usepackage[accepted]{icml2019}

\icmltitlerunning{Gradio: Hassle-Free Sharing and Testing of ML Models in the Wild}

\begin{document}

\twocolumn[
\icmltitle{Gradio: Hassle-Free Sharing and Testing of ML Models in the Wild}



\icmlsetsymbol{equal}{*}

\begin{icmlauthorlist}
\icmlauthor{Abubakar Abid}{equal,stanford_ee,gradio}
\icmlauthor{Ali Abdalla}{equal,gradio}
\icmlauthor{Ali Abid}{equal,gradio}
\icmlauthor{Dawood Khan}{equal,gradio}
\icmlauthor{Abdulrahman Alfozan}{gradio}
\icmlauthor{James Zou}{stanford_bio}
\end{icmlauthorlist}

\icmlaffiliation{stanford_ee}{Department of Electrical Engineering, Stanford University, Stanford, California, USA}
\icmlaffiliation{gradio}{Gradio Inc., Mountain View, California, USA}
\icmlaffiliation{stanford_bio}{Department of Biomedical Data Science, Stanford University, Stanford, California, USA}

\icmlcorrespondingauthor{Abubakar Abid}{a12d@stanford.edu}

\icmlkeywords{Machine Learning, ICML}

\vskip 0.3in
]



\printAffiliationsAndNotice{\icmlEqualContribution} 

\begin{abstract}
Accessibility is a major challenge of machine learning (ML). Typical ML models are built by specialists and require specialized hardware/software as well as ML experience to validate. This makes it challenging for non-technical collaborators and endpoint users (e.g. physicians) to easily provide feedback on model development and to gain trust in ML.  
The accessibility challenge also makes   collaboration more difficult and limits the ML researcher's exposure to realistic data and scenarios that occur in the wild.
 To improve accessibility and facilitate collaboration, we developed an open-source Python package, Gradio, which allows researchers to rapidly generate a visual interface for their ML models. Gradio makes accessing any ML model as easy as sharing a URL. Our development of Gradio is informed by interviews with a number of machine learning researchers who participate in interdisciplinary collaborations. Their feedback identified that Gradio should support a variety of interfaces and frameworks, allow for easy sharing of the interface, allow for input manipulation and interactive inference by the domain expert, as well as allow embedding the interface in iPython notebooks. We developed these features and carried out a case study to understand Gradio's usefulness and usability in the setting of a machine learning collaboration between a researcher and a cardiologist. 
\end{abstract}

\section{Introduction}
\label{section:introduction}

Machine learning (ML) researchers are increasingly part of interdisciplinary collaborations in which they work closely with domain experts, such as  doctors, physicists, geneticists, and artists \citep{bhardwaj2017study, radovic2018machine, zou2018primer, hertzmann2018can}. In a typical work flow, the domain experts will provide the data sets that the ML researcher analyzes, and will provide high-level feedback on the progress of a project. However, the domain expert is  usually very limited in their ability to provide direct feedback on model performance, since, without a background in ML or coding, they are unable to try out the ML models during development.

This causes several problems during the course of the collaboration. First, the lack of an accessible model for collaborators makes it very difficult for domain experts to understand when a model is working well and communicate relevant feedback to improve model performance. Second, it makes it difficult to build models that will be reliable when deployed in the real world, since they were only only trained on a fixed dataset and not tested with the domain shifts present in the real world (``in the wild"). Real-world data often includes artifacts that are not present in fixed training data; domain experts are usually aware of such artifacts and if they could access the model, they may be able to expose the model to such data, and gather additional data as needed \citep{thiagarajan2018can}. Lack of end-user engagement in model testing can lead to models that are biased or particularly inaccurate on certain kinds of samples. Finally, end-user domain experts who have not engaged with the model as it was being developed tend to exhibit a general distrust of the model when it is deployed. 

In order to address these issues, we have developed an open-source python package, Gradio\footnote{The name is an abbrevation of \textbf{grad}ient \textbf{i}nput \textbf{o}utput}, which allows researchers to rapidly generate a web-based visual interface for their ML models. This visual interface allows domain experts to interact with the model without writing any code. The package includes a library of common interfaces to support a wide variety of models, e.g. image, audio, and text-based models. Additionally, Gradio makes it easy for researchers to securely share their public links to their models so that collaborators can try out the model directly from their browsers without downloading any software, and also lets the domain expert provide feedback on individual samples,  fully enabling the feedback loop between domain experts and ML researchers. 

\begin{figure*}
    \centering
    \includegraphics[width=0.9\linewidth]{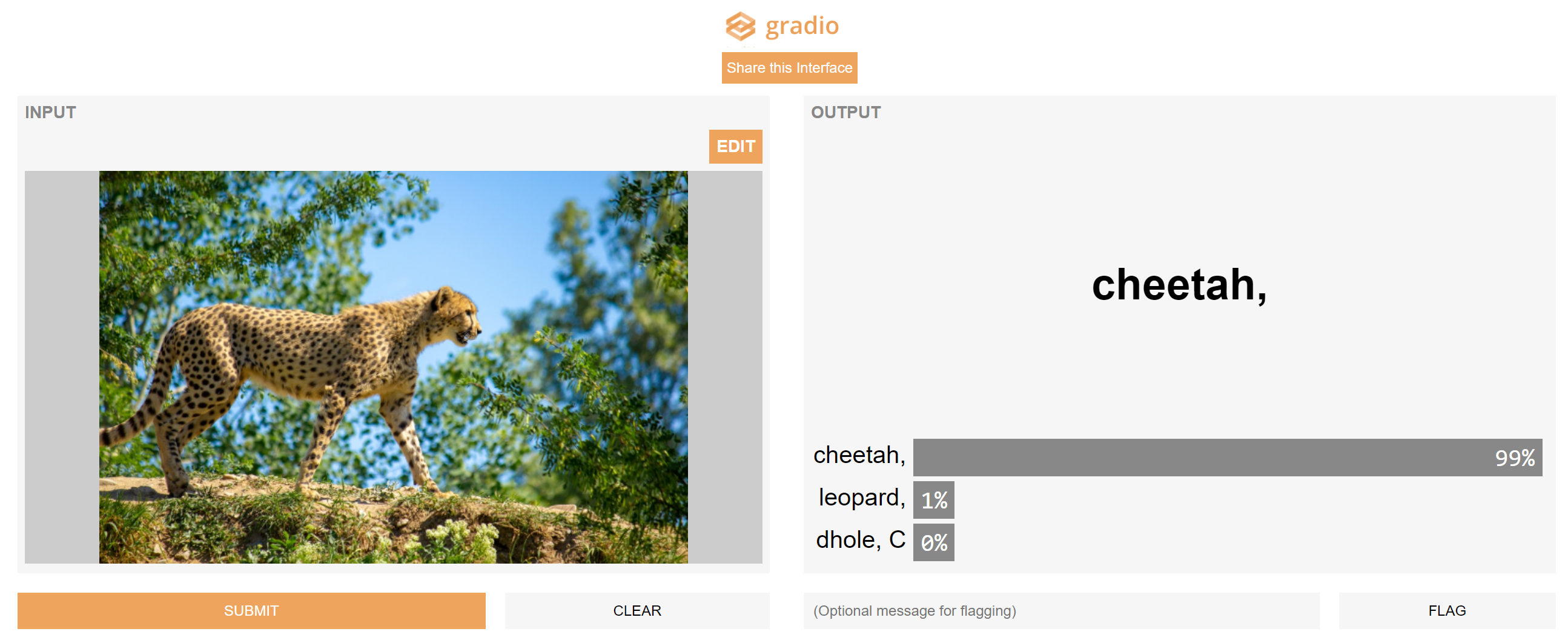}
    \caption{An illustration of a web interface generated by Gradio, which allows users to drag and drop their own images (left), and get predicted labels (right). Gradio can provide an interface wrapper around any machine learning model (InceptionNetv3 is shown in the example here). The web interface can be shared with others using the share link button (center top), and collaborators can provide feedback by flagging particular input samples (bottom right).}
    \label{fig:my_label}
\end{figure*}

In the rest of this paper, we begin by discussing related works and their limitations, which led to the development of Gradio (Section 2). We then detail the implementation of Gradio in Section 3. We have carried out a preliminary pilot study that includes an ML researcher and a clinical collaborator, which we describe in Section 4. We conclude with a discussion of the next steps for Gradio in Section 5.

\section{Motivation}

\subsection{Related Works}

The usefulness of visual interfaces in interdisciplinary collaborations has been observed by many prior researchers, who have typically created highly customized tools for specific use cases. For example \citet{xu2018ecglens} created an interactive dashboard to visualize ECG data and classify heart beats. The authors found that the visualization significantly increased adoption of the ML method and improved clinical effectiveness at detecting arrhythmias. 

However, visual interfaces that have been developed by prior researchers have been tightly restricted to a specific machine learning framework \citep{klemm2018barista} or to a specific application domain \citep{muthukrishna2019dash}. When we interviewed our users with regards to such tools, they indicated that the limited scope of these tools would make them unsuitable for their particular work. 

\subsection{Design Requirements}
\label{subsection:design_requirements}
We interviewed 12 machine learning researchers who participate in interdisciplinary collaborations. Based on the feedback gathered during these interviews, we identified the following key design requirements. 

\textbf{R1: Support a variety of interfaces and frameworks.} Our users reported working with different kinds of models, where the input (or output) could be: text, image, and even audio. To support the majority of models, Gradio must be able to offer developers a range of interfaces to match their needs. Each of these interfaces must be intuitive enough so that domain users can use them without a background in machine learning. In addition, ML researchers did not want to be restricted in which ML framework to use: Gradio needed to work with at least Scikit-Learn, TensorFlow, and PyTorch models.

\textbf{R2: Easily share a machine learning model.} Our users indicated that deploying a model so that it can be used by domain experts is very difficult. They said that Gradio should allow developers to easily create a link that can be shared with researchers, domain experts, and peers, ideally without having to package the model in a particular way or having to upload it to a hosting server.

\textbf{R3: Manipulate input data.} To support exploration and improvement of models, the domain expert needs the ability to manipulate the input. For example, the ability to crop an image, occlude certain parts of the image, edit the text, add noise to an audio recording, or trim a video clip. This helps the domain expert detect which features affect the model, and what kind of additional data needs to be collected in order to increase the robustness of the model.

\textbf{R4: Running in iPython notebooks \& embedding.} Finally, our users asked that the interfaces be run from and embedded in Jupyter and Google's Colab notebooks, as well as embedded in websites. A use case for many of our researchers was to expose machine learning models publicly after being trained. This would allow their models to be tested by many people e.g. in a citizen data science effort, or as part of a tutorial. They needed Gradio to both allow sharing of models directly with collaborators, but also widely with the general public.

\section{Implementation}
\label{subsection:implementation}

\begin{figure*}
    \centering
    \includegraphics[width=\linewidth]{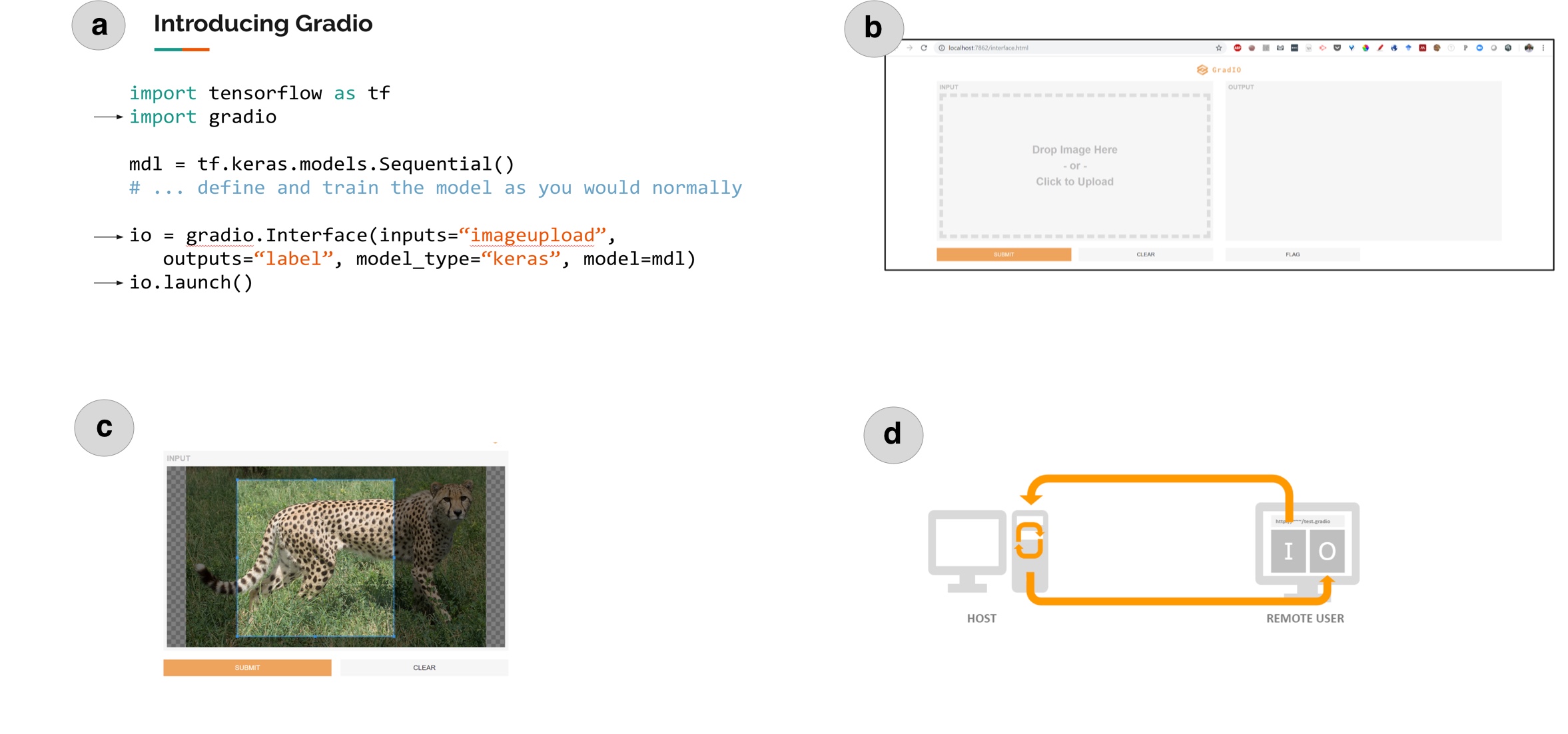}
    \caption{A diagram of the steps to share a machine learning model using Gradio. Steps: (a) The machine learning researcher defines the input and output interface types, and launches the interface either inline or in a new browser tab. (b) The interfaces launches, and optionally, a public link is created that allows remote collaborators to input their own data into the model. (c) The users of the interface can also manipulate the model in natural ways, such as cropping images or obscuring parts of the image. (d) All of the model computation is done by the host (i.e. the computer that called Gradio). The collaborator or user can interact with the model on their browser without local computation, and can provide real-time feedback (e.g. flagging incorrect answers) which is sent to the host. }
    \label{fig:steps}
\end{figure*}

Gradio is implemented as a python library, and can be installed from PyPi\footnote{\texttt{pip install Gradio}}. Once installed, running a Gradio interface requires minimal change to a ML developer's existing workflow. After the model is trained, the developer creates an \textit{Interface} object with four required parameters (Fig. \ref{fig:steps}a). The first and second parameter are \textit{inputs} and \textit{outputs}, which takes as argument the input/output interface to be used. The developer can choose any of the subclasses of \textit{Gradio.AbstractInput} and \textit{Gradio.AbstractOutput}, respectively. Currently this includes a library of standard interfaces for handling image, text, and audio data. The next parameter is \textit{model\_type} which is a string representing the type of model being passed in; This may be \textit{keras}, \textit{pytorch}, or \textit{sklearn} -- or it may be \textit{pyfunc}, which handles arbitrary python functions.  The final parameter is \textit{model} where the developer passes in the actual model to use for processing. Due to the common practice of pre-processing or post-processing the input and output of a specific model, we implemented a feature to instantiate \textit{Gradio.Input}/\textit{Gradio.Output} objects with custom parameters or alternatively supply custom pre-processing and post-processing functions.

We give the developer the option of how the interface should be launched. The launch function accepts 4 boolean variables, which allow for displaying the model \textit{inbrowser} (whether to display model in a new browser window) (Fig. \ref{fig:steps}b), displaying it \textit{inline} (whether to display model embedded in interactive python environment such as Jupyter or Colab notebooks), attempting to \textit{validate} (whether to validate the interface-model compatibility before launching), and creating a  \textit{share} link (whether to create a  public link to the model interface). If Gradio creates a share link to the model, then the model continues running on the host machine, and an SSH tunnel is created allowing collaborators to pass in data into the model remotely, and observe the output. This allows the user to continue running using the same machine, with the same hardware and software dependencies. The collaborator does not need any specialized hardware or software: just a browser running on a computer or mobile phone (the user interfaces are mobile-friendly).

The user of the interface can input any data and also manipulate the input by, for example, cropping an image (Fig. \ref{fig:steps}c). The data from the input is encrypted and then is passed securely through the SSH tunnel to the developer's computer, which is actually running the model, and the output is passed back to the end user to display (\url{www.gradio.app} also serves as a coordinator service between public links and the SSH tunnels). The amount of time until the end user receives the output is simply the amount of time it takes for model inference plus any network latency in sending data. The collaborator can additionally flag data where the output was false, which sends the inputs and outputs (along with a message) to the ML researcher's computer, closing the feedback loop between researcher and domain expert. (Fig. \ref{fig:steps}d).

\section{Pilot Study: Echocardiogram Classification}
\label{subsection:pilot_study}

\begin{figure*}[t!]
    \centering
    \begin{subfigure}[t]{\textwidth}
        \centering
        \includegraphics[width=\textwidth]{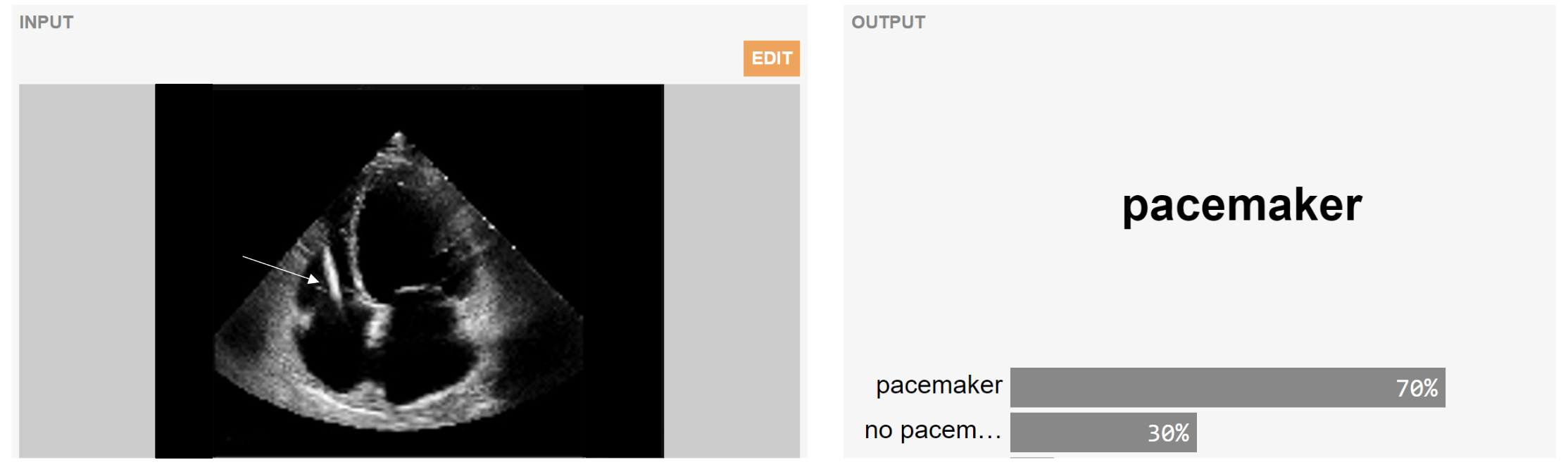}
        \caption{}
    \end{subfigure}\\
    \begin{subfigure}[t]{\textwidth}
        \centering
        \includegraphics[width=\textwidth]{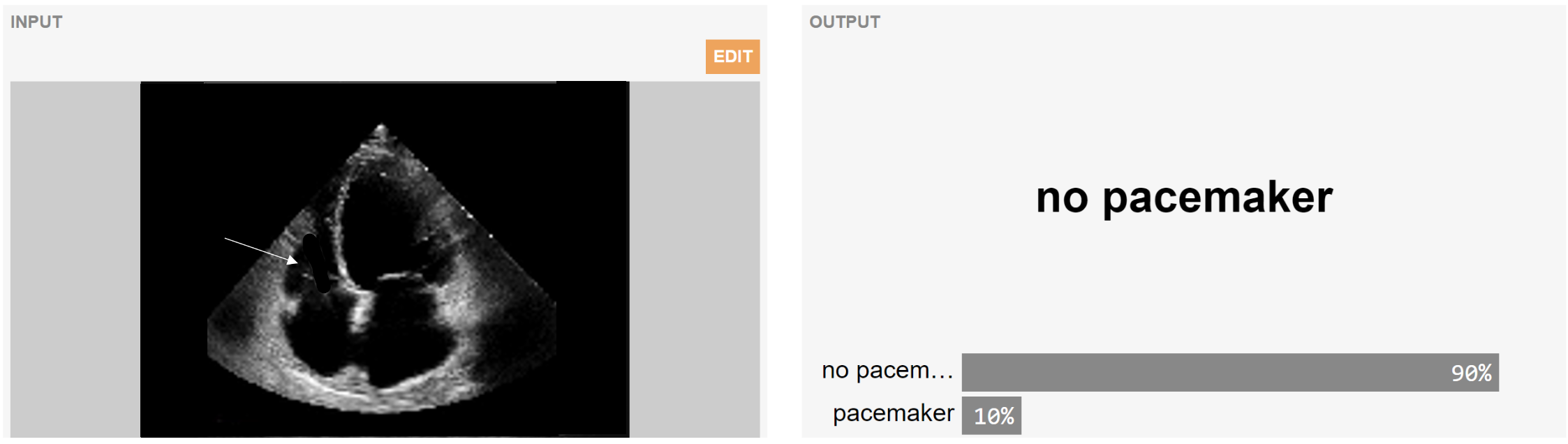}
        \caption{}
    \end{subfigure}
    \caption{In our pilot study, the clinician used Gradio to test a model that classified echocardiograms based on the presence of a pacemaker. (a) The clinician submitted his own image of an echocardiagram, similar to the one shown here, and the model correctly predicted that a pacemaker was present. (b) The clinician used Gradio's built-in tools to obscure the pacemaker and the model correctly predicted the absence of a pacemaker. (The white arrows are included to point out the location of the pacemaker to the reader; they were not present in the original images).}
    \label{fig:pacemaker}
\end{figure*}

We carried out a user study to understand the usefulness of \lstinline{Gradio} in the setting of an ML collaboration. The participants were an ML researcher and a cardiologist who had in collaboration developed an ultrasound classification model that could determine whether a pacemaker was present in a patient from a single frame in an ultrasound video. The model scored an area under receiver-operating characteristic curve (AUC) of 0.93 on the binary classification task.

We first asked participants a series of questions to record their typical workflow without using the Gradio library. We then taught the participants the Gradio library and let them use it for collaborative work. After being shown instructions about the Gradio library, the ML researcher was able to set up Gradio on a lab server that was running his model. The process of setting up Gradio took about 10 minutes, as some additional python dependencies needed to be installed on the lab server. After the installation, the researcher was able to copy and adapt the standard code from Gradio documentation and did not run into any bugs. The ML researcher was advised to share the model with the cardiologist. He did so, and the cardiologist automatically began to test the robustness of the model by inputting his own images and observing the model's response. After using Gradio, the cardiologist gained more confidence and trust in the performance of this particular ML model.

We observed the researchers while they carried out these tasks, as we sought to answer four research questions:

\textbf{Q1: How do researchers share data and models with and without Gradio?}

Before Gradio, the cardiologist provided the entire dataset of videos to the machine learning researcher in monthly batches. These batches were the latest set of videos that were available to the cardiologist. The ML researcher would train the model on increasingly larger datasets and report metrics such as classification accuracy and AUC to the cardiologist. Beyond this, there was very little data sharing from the cardiologist and the researcher did not ever share the model with the cardiologist.

With Gradio, the cardiologist opened the link to the model sent by the ML researcher. Even though it was his first time using the model, the cardiologist immediately began to probe the model by inputting an ultrasound image from his desktop into the Gradio model. He chose an image which clearly contained a pacemaker, see Fig. \ref{fig:pacemaker}(a). The model correctly predicted that a pacemaker was present in the patient. The cardiologist then occluded the pacemaker using the paint tool built into Gradio, see Fig. \ref{fig:pacemaker}(b). After completely occluding the pacemaker, the cardiologist resubmitted the image; the model switched its prediction to ``no pacemaker,'' which elicited an audible sigh of relief from the ML researcher and cardiologist.

The cardiologist proceeded to choose more difficult images, generally finding that the model correctly determined when a pacemaker was and was not present in the image. He also occluded different regions in the image to serve as a comparison to occluding the pacemaker. The model performance was generally found to be accurate and robust; a notable exception was in the case of flipping over the vertical axis, which would generally significantly affect the prediction accuracy. When this happened, the cardiologist flagged the problematic images, sending them to the ML researcher's computer for further analysis.

\textbf{Q2: What features of Gradio are most used and most unused by developers and collaborators?}

We found that the machine learning researcher quickly understood the different interfaces available to him for his model. He selected the appropriate interface for his model, and set \textit{share=True} to generate a unique publicly accessible link for the model. When he shared the link with the cardiologist, the cardiologist spent a great deal of time trying different ways to manipulate a few sample images to affect the model prediction. The cardiologist treated this as a challenge and tried to cause the model to make a mistake in an adversarial manner. The cardiologist also used the flagging feature in the cases where the model did make a mistake. 

The ``share'' button which appears at the top of the interface was not used; instead, the ML researcher simply copied and pasted the URL to share it with his collaborator. And despite the general interest in running  model interfaces inside iPython notebooks, our users did not use that feature. 

\textbf{Q3: What kind of model feedback do collaborators provide to developers through the Gradio interface?}

The collaborator tested various transformations on the test images, including changing the orientation of the image to occluding parts of the image. Whenever this would cause the model to make a mistake on an image, the cardiologist would flag the image, but would usually pass a blank message. Thus, it seemed that the collaborator would only send images that were misclassified back to the ML researcher.

\textbf{Q4: What additional features are requested by the developers and collaborators?}

Our users verbally requested two features as they were using the model. First, the cardiologist asked if it would be possible for the ML developer to pre-supply images to the interface. This way, he would not need to find an ultrasound image from his computer, but would be able to choose one from a set of images already displayed to him.

Second, the collaborator was used to seeing saliency maps for ultrasound images that the ML researcher had generated in previous updates. The collaborator expressed that it would be very helpful for him to see these saliency maps, especially as he was choosing what areas inside of the image to occlude.

\section{Discussion \& Next Steps}
\label{subsection:discussion}

In this paper, we describe a Python package that allows machine learning researchers to easily create visual interfaces for their machine learning models, and share them with collaborators. Collaborators are then able to interact with the machine learning models without writing code, and provide feedback to the machine learning researchers. In this way, collaborators and end users can test machine learning models in settings that are realistic and can provide new data to build models that work reliably in the wild.

We think this will lower the barrier of accessibility for domain experts to use machine learning and take a stronger part in the development cycle of models. At a time when machine learning is becoming more and more ubiquitous, the barrier to accessibility is still very high.

We carried out a case study to evaluate the usability and usefulness of Gradio within an existing collaboration between an ML researcher and a cardiologist working on detecting pacemakers in ultrasounds. We were surprised to see that both the ML researcher and the domain expert seemed to be relieved when the model worked, as though they expected it not to. We think because researchers can not manipulate inputs the way a domain expert would, they generally have less confidence in the model's robustness in the wild. At the same time, because domain experts have not interacted with the model or used it, they feel the same doubt in its robustness. This study was however limited in scope to one pair of users, and a short time. We plan to conduct more usability studies and quantitative measures of trust in machine learning models to get a better holistic view of the usability and usefulness of Gradio. Similarly, quantitative measures of user satisfaction on both the end of machine learning researcher and domain expert can be used to evaluate the product and guide its further development.

The next steps in the development of the package would be creating features for saliency, handling other types of inputs (ex: tabular data), handling bulk inputs, as well as helping ML researchers reach domain experts even if they don't already have access to them. 

Additional documentation about Gradio and example code can be found at: \url{www.gradio.app}.

\section*{Acknowledgments}
\label{section:acknowledgments}

We thank all of the machine learning researchers who talked with us to help us understand the current difficulties in sharing machine learning models with collaborators, and gave us feedback during the development of Gradio. In particular, we thank Amirata Ghorbani and David Ouyang for participating in our pilot study and for sharing their echocardiogram models using Gradio. 

\bibliography{main}
\bibliographystyle{icml2019}

\end{document}